\pdfoutput=1

\documentclass[11pt]{article}

\usepackage{acl}

\usepackage{times}
\usepackage{latexsym}

\usepackage[T1]{fontenc}

\usepackage[utf8]{inputenc}

\usepackage{microtype}

\usepackage{graphicx}
\usepackage{color,soul}
\usepackage{multirow}

\title{Regex in a Time of Deep Learning: The Role of an Old Technology in Age Discrimination Detection in Job Advertisements}

\author{Anna Pillar\textsuperscript{1,2}, Kyrill Poelmans\textsuperscript{2}, Martha Larson\textsuperscript{1} \\
\textsuperscript{1}Radboud University, Netherlands \\ \textsuperscript{2}Textmetrics, Netherlands \\
\texttt{anna@textmetrics.com, kyrill@textmetrics.com,} \\
\texttt{martha.larson@ru.nl}}

\begin{document}
\maketitle

\begin{abstract}
Deep learning holds great promise for detecting discriminatory language in the public sphere.
However, for the detection of illegal age discrimination in job advertisements, regex approaches are still strong performers.
In this paper, we investigate job advertisements in the Netherlands.
We present a qualitative analysis of the benefits of the `old' approach based on regexes and investigate how neural embeddings could address its limitations.
\end{abstract}

\section{Introduction}
Age discrimination is often related to work and it starts in the pre-hiring phase with job advertisements.
Each year, thousands of job descriptions in the Netherlands contain age discrimination, which is illegal under Dutch law~\cite{Fokkens2018}.

The state of the art in detection of illegal age discrimination in Dutch job ads uses regular expressions (regex)~\cite{Fokkens2018}.
This `old' approach works surprisingly well because illegal age discrimination uses predictable vocabulary, and keywords such as `age' are quite reliable indicators.
However, individual sentences from job ads suggest that neural embedding approaches, with their ability to capture semantics, could also be helpful, e.g., `Given our own advancing years, it would be just lovely to have a younger soul join us.'

The contribution of this paper is a qualitative analysis of the role that regex should continue to play in  detecting illegal age discrimination, now that the language technology community has moved towards deep learning approaches. 
Since regexes offer explainable decisions, we do not seek to abandon the regex approach, but rather to understand its potential compared with the potential of neural embeddings.
Because it is known that the regex approach can suffer from low recall~\cite{Fokkens2018}, our main focus is on understanding false positives (i.e., cases of discrimination that the detector misses).



In this paper, we report the essential findings on illegal age discrimination detection in Dutch job ads of a larger study~\cite{pillar}, which contains further analysis.
After a brief introduction to age discrimination (Sec.~\ref{background}) and the regex approach of ~\citet{Fokkens2018} (Sec.~\ref{related}), we present two analyses.
The first (Sec.~\ref{shortcomings}) investigates the regex approach, which is currently the state of the art.
The second (Sec.~\ref{neural}) looks at whether and how neural embeddings could complement regexes in the future.

Our analyses make use of the Job Digger dataset, which contains 1.2 million Dutch job advertisements collected by a Dutch company, Job Digger, and made available to us for use in our study.
Job Digger had created the dataset by carrying out a large scale crawl of internet job postings in the Netherlands in 2014.
The comprehensiveness of this crawl ensures that our dataset is representative of the full spectrum of possible Dutch job ads.

Our investigation reveals that the regex approach is more difficult to improve upon than one might think. 
The final section of the paper (Sec.~\ref{Conclusion}) provides an outlook and discusses how researchers in the future should seek to leverage both regexes and neural embeddings for explainable detection of illegal age discrimination.

\section{Background and Related Work}\label{background}
Age discrimination is defined as bias and prejudice against people based on their age and ageism is 
one of the three big `isms', next to sexism and racism~\cite{Butler1969}.
In practice, age discrimination predominantly targets older people~\cite{Bytheway2005}. 
Ageism is in this sense unique among `isms' because, in the natural course of life, in-group members become out-group members \cite{Jonson2013WeAge}.
However, despite the fact that it threatens everyone, ageism is difficult to fight. 
It is culturally acceptable~\cite{Gendron} and people are unaware of it~\citep{Palmore2001TheSurvey}.
In the Netherlands, concern about age discrimination has grown recently, mainly in employment~\cite{Andriessen2014ErvarenNederland}.

 

Age discrimination occurs in two main forms~\citep{Voss2018}. 
\emph{Objective Ageism} is defined through legal frameworks that protect the vulnerable group from discrimination.
\emph{Subjective Ageism} (or \emph{Perceived Ageism}) 
is bias and discrimination that does not fall under a legal definition.

In the Netherlands, 
the Dutch Equal Treatment Act regarding age discrimination at the workplace 
prohibits discrimination in the context of work, including job advertisements.
The law defines two forms of discrimination:
\textit{Direct discrimination} involves an explicit mention of the age of the candidate, e.g., `You are younger than 30 years'. 
\textit{Indirect discrimination}, 
involves formulations that imply age, e.g., specifically recruit students (who, in the Netherlands, characteristically are young). 

The literature 
on age discrimination detection in job ads is surprisingly limited.
The work closest to ours studied the relationship between stereotypes in English-language job ads and in hiring~\cite{Burn2019}.
It implemented an age discrimination detector for job ads, but focused on stereotypes, which are not necessarily illegal.
In contrast, we study detection of discriminatory statements that are explicitly defined, and prohibited, by law.


\section{Regex Baseline}
\label{related}

The state of the art in the detection of age discrimination in Dutch job ads~\cite{Fokkens2018} uses a list of keywords to detect objective ageism.
The keywords were identified by manually reading a large number of job ads. 
They were selected because they were judged to be indicative of illegal discrimination when used in certain contexts.

Appendix A
contains the keyword list with a sample sentence from a job ad for each keyword.
The keywords form the basis of a set of regular expressions, which~\citet{Fokkens2018} constructed with the aim of covering all possible contexts in which each keyword could be discriminatory.
The importance of context is illustrated by the following example.
The sentences, `You will be responsible for young students' contains both the words `young' and `student', but is not discriminatory because the words describe the job and not the candidate.
\citet{Fokkens2018} published a set of these regexes on GitHub\footnote{\url{https://github.com/cltl/AgeDiscriminationBaseline}}.
They discovered that regexes perform best if they allow a certain amount of flexibility by including the white card character \texttt{.\{0,30\}}, e.g., `\texttt{you\textbackslash s+are\textbackslash s+a\textbackslash s+.\{0,30\}student}'.
They report that such flexible regexes achieve a high precision (94.5\%), but a somewhat low recall (75.7\%) on their test set. 

\section{Role of the Regex Baseline}
\label{shortcomings}
In this section, we discuss our first qualitative analysis, which aimed to reveal both the potential and the inherent weaknesses of the regex approach.

\subsection{Data and Annotation}
We created a representative dataset large enough to yield interesting insights but small enough to be hand annotated by sampling ca. 3,000 
sentences from the Job Digger dataset.
About half of the sentences we sampled were selected to contain one keyword, but to not match any regexes.
The inclusion of a large number of these sentences improved the chance that we could gain insight into how the inherent weaknesses of regexes might contribute to false positives.
We consider a weakness `inherent' if it relates to expressiveness or generalizabilty of the regexes themselves, rather than to the exact keywords we are using.
As much as possible, we sampled evenly over the keywords.
About a third of our sample sentences were chosen to match a regex.
The samples in the remaining ca. 10\% of the dataset did not include a keyword.


The data set was annotated for age discrimination by a group of seven annotators with good familiarity with Dutch law, who were split into two teams.
Each sample was annotated by two annotators, one from each team.
The inter-annotator agreement (Cohen's Kappa) 
between teams reflected substantial agreement ($\kappa=0.61$).
Samples on which the annotators disagreed or where one was unsure were not included in our dataset, leaving a total of 2,195 annotated samples for analysis.
 
\subsection{Approach and Findings}
We conducted our analysis by inspecting sample sentences by hand and investigating two levels: (1) at a general level across all keywords (2) at a keyword level, focused on the false negatives associated with each keyword.
We report our findings organized into a set of insights:
\\
\\
\\

\textbf{General sentence length and structure}
Across the keywords, we found variation in sentence length and structural complexity, from bullet points such as `- Age up to 27 years' to verbose sentences such as `We are looking for man and especially also for women, who know the shop floor inside out, and are between 50 and 70 years of age.'
The regexes in our list were too elaborate to capture the bullets and too narrow to capture the verbose sentences.
This observation points to an inherent limitation of regexes.
Our analysis also revealed a certain number of frequent formulation for which a regex missing keywords or a missing formulation could easily be added. 




\textbf{Keyword-specific issues}
When looking at the sample sentences of individual keywords 
we found that the issue of sentence length and structure occurred across keywords, but was a particular issue for certain keywords, specifically,
`young' (\textit{jong}) and `age' \textit{(leeftijd}). 
This observation suggests that not all keywords should be handled the same.

\textbf{Keyword context}
At the keyword level, we found that for `young' (\textit{jong}) and  `recent graduate' (\textit{schoolverlater)}, the discrimination is determined by the context in which they are used.
 As mentioned above, if these keywords are used to describe the job and not the candidate, they are not discriminatory.
We found that the formulations used were very open.
There seemed to be no frequent formulation that could be added to the regexes to cover the variety of the samples in which the context was not captured by the regexes, causing a false negative.

\textbf{Keywords associated with discrimination}
We observed that some keywords seem to be associated with discrimination, but did not themselves directly express discrimination.
For example, the keyword `extra money' (\textit{bijverdienen}) as used in the sentence `Have you recently completed your degree and would like to earn a earn a little extra money?' is not causing the sentence to be discriminatory.
Rather, the reference to `recent graduation' makes the sentence discriminatory.
This observation suggests that better modeling of context can improve the performance of regexes.

\textbf{Limited non-discriminatory usage} 
Certain keywords, such as, e.g., `recent graduate', just discussed, mainly occur in discriminatory sentences. 
However, in 3 out of 114 samples with the keyword `recent graduate', it was actually used in a non-discriminatory way. 
This observation suggests that regexes should be designed to capture the non-discriminatory contexts.
If a sentence containing a keyword does not match a `non-discriminatory regex' then it can be considered discriminatory.

\section{Role of Neural Embeddings}
\label{neural}
In this section, we discuss our second qualitative analysis, which aimed to discover how neural embeddings can potentially complement regex.

Since the issue of missing keywords was already raised by~\citet{Fokkens2018}, we focus on another property of regexes that Sec.~\ref{shortcomings} revealed to be an issue for detection of illegal age discrimination: they cannot capture discrimination when it is phrased using different syntax but expresses similar semantics.
This inflexibility becomes particularly important when we consider the importance of modeling the broader context of a keyword within a sentence.


\subsection{Approach and findings}
Our analysis consisted of manual inspection of a large number of sentence embedding clusters.
We trained ALBERT word embeddings~\cite{LanALBERT:REPRESENTATIONS} on ~5 million sentences drawn from the Job Digger dataset. 
The training was done from scratch with the MLM learning task.
To create sentence embeddings, we averaged the word embeddings of the component words, following common practice.

Our hope was that in the sentence embedding space, we would observe a separation between discriminatory and non-discriminatory sentences, since these express different semantics. 
However, when we visualized our samples using t-SNE~\cite{VanDerMaaten2008VisualizingT-SNE}, we did not observe clear discriminating and non-discriminatory clusters.
We concluded that a standardly trained semantic space cannot easily capture age discrimination and turned to analyze if neural embeddings could capture useful differences in keyword context.

\begin{figure}[h]
    \centering
    \includegraphics[width=0.5\textwidth]{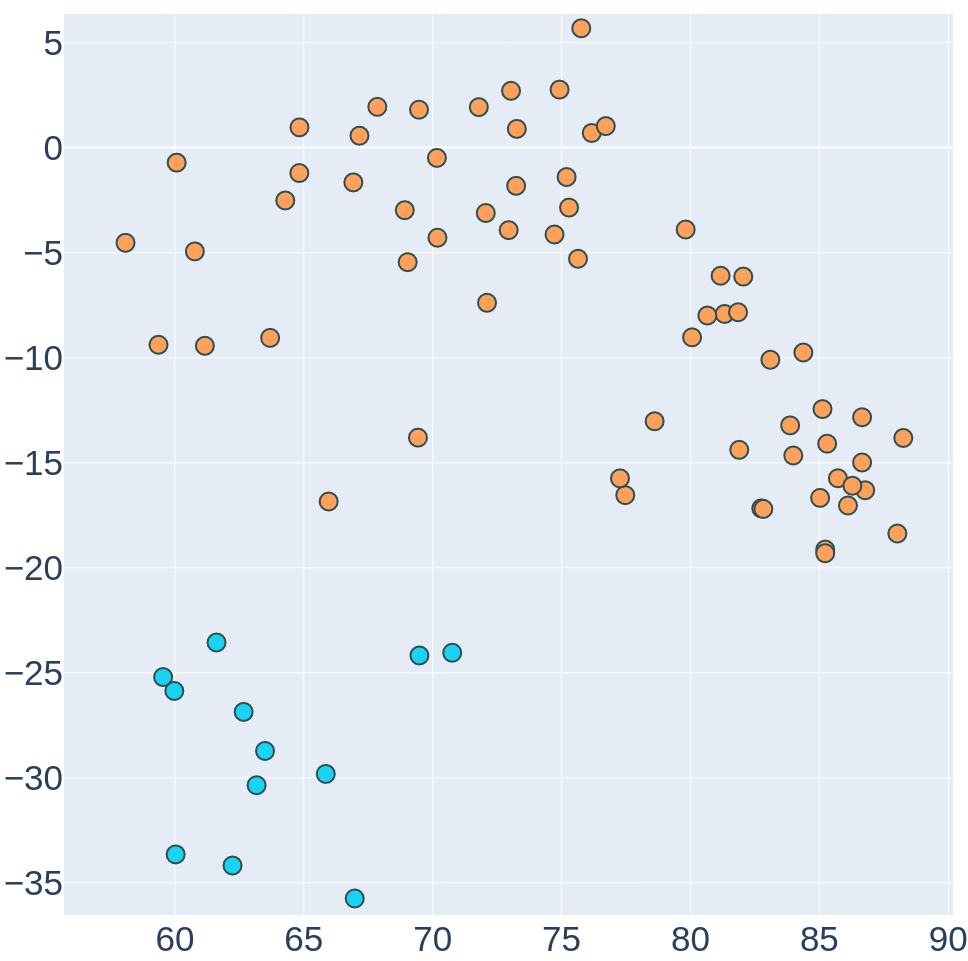}
    \caption{An excerpt of the plot of the embeddings of sentences containing the keyword `between'. Orange: discriminatory, Blue: non-discriminatory}
    \label{fig:tussen}
\end{figure}

For each keyword, we selected the sentences in our annotated data set that contained it and visualized them with t-SNE.
In most cases, the discriminatory and non-discriminatory sentences were not well separated.
However, there were a few cases that are worth further discussion\footnote{Full interactive plots for all keywords can be found at \small{\url{https://github.com/Textmetricslab/Regex-in-a-Time-of-Deep-Learning}}}.

\textbf{Keyword `between'} 
Good separation was observed for the keyword `between', as can be seen in Fig. \ref{fig:tussen}. 
The discriminatory sentences include: `You are preferably aged between 17 and 20 years.' and `Are you the person we are looking for and are you are aged between 16 and 19 years?'
The cluster in the lower part of the plot consists of samples that use the word `between' to give information about the work time and are not discriminatory: `You will be working between 10 and 25 hours per week, from Monday to Sunday' and `Total work time per week is between 8 and 12 hours'.

It is interesting to note that all discriminatory samples contain a number followed by the word `age' and  non-discriminatory samples contain a number followed by either `hour' or its abbreviation. 
This means that in this case, regexes could have also distinguished these two contexts.

\textbf{Keyword `experience'} Another interesting example was the keyword  `experience', which is discriminatory if it limits the years of experience (e.g., `you have a maximum of 5 years of experience'), but not if it specifies the minimum years of experience needed.
When we visualized the sentences containing the keyword `experience', we observed no separation between these two cases.
However, we did see a cluster of non-discriminatory samples that all stated that salary would be based on experience, which is non-discriminatory. 
Possibly, regexes based `salary'-related keywords also could capture the difference between these contexts.

\textbf{Keyword `old'} The keyword `old' also yielded an interesting observation.
A cluster of sentences containing `old' all directly address candidates and mentioned an desired age, e.g.,:
`Are you enthusiastic, like to (physically) work and are you between 18 and 30 years old?'; `You are minimally 23 years old.'; and
`Are you between 18 and 26 years old?'.
However, the cluster also contained the sentence `Are you badass commercial, entrepreneurial, a builder, mobile, never too old to learn, do you go for freedom, are you studious, is hierarchy something you are allergic for and are you often smarter than your boss?'.
It fits the general style of directly addressing the candidate (`Are you...`) and also contains the word `old'. 
However, the usage of `old' in this context is not discriminatory but rather part of a description of the candidates attitude.

In sum, our qualitative analysis leads us to conclude that neural embeddings do not offer a silver-bullet solution to improving detection of illegal age discrimination over what is already possible using regexes.
We did not uncover evidence that suggests that it would be worthwhile to trade in the explainability of the regex approach for benefits offered by using sentence embeddings.

\section{Conclusion and Outlook}\label{Conclusion}
In this paper, we have investigated the contribution of regexes to the task of automatically detecting illegal age discrimination in Dutch job ads.
We have found there is potential to improve the recall of the regex lists of ~\citet{Fokkens2018}, which constitute the current state of the art, not only by adding keywords, but also by creating additional regexes.
 
Future work should investigate a simple approach based on rule mining, which was not explored by~\citet{Fokkens2018}.
In~\cite{pillar}, we report an exploration of automating the generation of regular expressions using active learning and genetic programming, but more work is necessary if these directions are to yield fruit.

The results of our analysis suggest that there is little to be gained in using neural embeddings directly in age discrimination detectors.
Instead, neural embeddings could have a role in the discover of new keywords and new regexes, extending a simple rule mining approach.
Using neural embeddings in this way would allow us to continue to benefit from the explainability of the regex approach.


The results of our qualitative study are not dependent on particular keywords, writing styles, or special properties of the Dutch language.
For this reason, we expect that our findings can be reproduced using other datasets and in other languages.
In fact, regex has been successfully used for general discrimination detection in Indonesian job ads~\cite{ningrum2020}.
Reproduction of our study will confirm and extend our findings, ensuring that the `old' technology of regex is not discarded for a task for which it is well suited.

    

\section{Acknowledgment}
We want to thank everyone involved in the annotation of the dataset used in this research.
Their work made our investigations possible.
\bibliography{bibtex_paper}
\bibliographystyle{acl_natbib}

\clearpage

\begin{table*}[]
\caption{(Appendix A) The list of discriminatory keywords from~\cite{Fokkens2018} used in our work, each illustrated with a sentence from our dataset that was annotated as discriminatory (translated from Dutch).}
\label{tab:age-disc samples}
\resizebox{\textwidth}{!}{%
\begin{tabular}{|cl|}
\hline
\multicolumn{2}{|c|}{\textbf{DIRECT DISCRIMINATION}}                                                                                                                                                                                                                                                        \\ \hline
\multicolumn{1}{|c|}{Keyword}                                                           & Sample Sentence                                                                                                                                                                                                   \\ \hline
\multicolumn{1}{|c|}{young}                                                             & \begin{tabular}[c]{@{}l@{}}For several companies in the Alkmaar region we are looking for young, \\ motivated candidates who can be deployed flexibly.\end{tabular}                                               \\ \hline
\multicolumn{1}{|c|}{\begin{tabular}[c]{@{}c@{}}young part \\ (of a team)\end{tabular}} & \begin{tabular}[c]{@{}l@{}}In this role you will be part of a young and dynamic team who are jointly \\ responsible for the design and realization of infrastructure projects up \\ to ± €6 Million.\end{tabular} \\ \hline
\multicolumn{1}{|c|}{fit into a young team}                                             & We work with a young team, where you will definitely fit in!                                                                                                                                                      \\ \hline
\multicolumn{1}{|c|}{\multirow{2}{*}{age}}                                              & Age range 20 - 25 years;                                                                                                                                                                                          \\ \cline{2-2} 
\multicolumn{1}{|c|}{}                                                                  & Given the age structure of our team, we prefer a young colleague.                                                                                                                                                 \\ \hline
\multicolumn{1}{|c|}{age from to}                                                       & \begin{tabular}[c]{@{}l@{}}We ask boys and girls aged 16 - 25 years who are full of energy and \\ like to promote this gym!\end{tabular}                                                                          \\ \hline
\multicolumn{1}{|c|}{age to}                                                            & \begin{tabular}[c]{@{}l@{}}Are you enthusiastic, eager to learn, entrepreneurial and in the age \\ group up to 22 years?\end{tabular}                                                                             \\ \hline
\multicolumn{1}{|c|}{age from}                                                          & Age from 30 years, we have a big preference for 45 +                                                                                                                                                              \\ \hline
\multicolumn{1}{|c|}{old}                                                               & \begin{tabular}[c]{@{}l@{}}Are you enthusiastic, do you like to work and are you between 18 \\ and 30 years old?\end{tabular}                                                                                     \\ \hline
\multicolumn{1}{|c|}{in-between}                                                        & You are between 18 and 25 years old;                                                                                                                                                                              \\ \hline
\multicolumn{1}{|c|}{at least}                                                          & We are looking for full-time hospitality professionals, at least 25 years old                                                                                                                                     \\ \hline
\multicolumn{2}{|c|}{\textbf{INDIRECT DISCRIMINATION}}                                                                                                                                                                                                                                                      \\ \hline
\multicolumn{1}{|c|}{job}                                                               & Are you a graduate looking for your first-ever job?                                                                                                                                                               \\ \hline
\multicolumn{1}{|c|}{side-job}                                                          & Are you looking for an interesting job in addition to your studies?                                                                                                                                               \\ \hline
\multicolumn{1}{|c|}{earn money}                                                        & \begin{tabular}[c]{@{}l@{}}Have you just finished school or just graduated and want to earn \\ some extra money before you go on vacation?\end{tabular}                                                  \\ \hline
\multicolumn{1}{|c|}{experience}                                                        & \begin{tabular}[c]{@{}l@{}}Experience: You have a college education and have 1 to 3 years of \\ working experience in the media and/or IT industry.\end{tabular}                                                  \\ \hline
\multicolumn{1}{|c|}{education}                                                         & You are following the HBO education Construction?                                                                                                                                                                 \\ \hline
\multicolumn{1}{|c|}{\begin{tabular}[c]{@{}c@{}}recent \\ graduate\end{tabular}}        & \begin{tabular}[c]{@{}l@{}}For one of our clients we are looking for serious, enthusiastic recent \\ graduated who want to be trained as logistics employees.\end{tabular}                                        \\ \hline
\multicolumn{1}{|c|}{\multirow{2}{*}{step}}                                             & Are you eager to learn and looking for the first step in your career?                                                                                                                                             \\ \cline{2-2} 
\multicolumn{1}{|c|}{}                                                                  & Are you ready for the second step in your career?                                                                                                                                                                 \\ \hline
\multicolumn{1}{|c|}{study}                                                             & \begin{tabular}[c]{@{}l@{}}This job is excellent to combine with your studies and is a great \\ addition to your CV!\end{tabular}                                                                                 \\ \hline
\multicolumn{1}{|c|}{start}                                                             & \begin{tabular}[c]{@{}l@{}}For our client, we are looking for an enthusiastic and spirited starter \\ for the position of Online Marketer.\end{tabular}                                                           \\ \hline
\multicolumn{1}{|c|}{lesson schedule}                                                   & With great regularity we have on-call jobs that fit perfectly with your class schedule.                                                                                                                           \\ \hline
\end{tabular}%
}
\end{table*}

\end{document}